\documentclass[sigconf]{acmart}

\DeclareMathOperator{\bE}{\mathbb{E}}

\usepackage{xcolor}
\usepackage[ruled,vlined]{algorithm2e}
\usepackage{amsmath}
\newtheorem{theorem}{Theorem}

\copyrightyear{2023}
\acmYear{2023}
\setcopyright{acmlicensed}\acmConference[WWW '23]{Proceedings of the ACM Web Conference 2023}{May 1--5, 2023}{Austin, TX, USA}
\acmBooktitle{Proceedings of the ACM Web Conference 2023 (WWW '23), May 1--5, 2023, Austin, TX, USA}
\acmPrice{15.00}
\acmDOI{10.1145/3543507.3583259}
\acmISBN{978-1-4503-9416-1/23/04}

\begin{document}

\title{Two-Stage Constrained Actor-Critic for Short Video Recommendation}

\author{Qingpeng Cai}
\affiliation{%
  \institution{Kuaishou Technology}
  \city{Beijing}
  \country{China}}
\email{caiqingpeng@kuaishou.com}

\author{Zhenghai Xue}
\affiliation{%
  \institution{Kuaishou Technology}
  \city{Beijing}
  \country{China}}
\email{xuezhenghai@kuaishou.com}

\author{Chi Zhang}
\affiliation{%
  \institution{Kuaishou Technology}
  \city{Beijing}
  \country{China}}
\email{zhangchi08@kuaishou.com}

\author{Wanqi Xue}
\affiliation{%
  \institution{Kuaishou Technology}
  \city{Beijing}
  \country{China}}
\email{xuewanqi@kuaishou.com}

\author{Shuchang Liu}
\affiliation{%
  \institution{Kuaishou Technology}
  \city{Beijing}
  \country{China}}
\email{liushuchang@kuaishou.com}

\author{Ruohan Zhan}
\affiliation{%
  \institution{Hong Kong University of Science and Technology}
  \city{Hong Kong}
  \country{China}}
\email{rhzhan@ust.hk}

\author{Xueliang Wang}
\affiliation{%
  \institution{Kuaishou Technology}
  \city{Beijing}
  \country{China}}
\email{wangxueliang03@kuaishou.com}

\author{Tianyou Zuo}
\affiliation{%
  \institution{Kuaishou Technology}
  \city{Beijing}
  \country{China}}
\email{zuotianyou@kuaishou.com}

\author{Wentao Xie}
\affiliation{%
  \institution{Kuaishou Technology}
  \city{Beijing}
  \country{China}}
\email{xiewentao@kuaishou.com}

\author{Dong Zheng}
\affiliation{%
  \institution{Kuaishou Technology}
  \city{Beijing}
  \country{China}}
\email{zhengdong@kuaishou.com}

\author{Peng Jiang}
\affiliation{%
  \institution{Kuaishou Technology}
  \city{Beijing}
  \country{China}}
\email{jiangpeng@kuaishou.com}
\authornote{Corresponding author}

\author{Kun Gai}
\affiliation{%
  \institution{Unaffiliated}
  \city{Beijing}
  \country{China}}
\email{gai.kun@qq.com}

\renewcommand{\shortauthors}{Cai, et al.}




\begin{abstract}
  The wide popularity of short videos on social media poses new opportunities and challenges to optimize recommender systems on the video-sharing platforms. Users sequentially interact with the system and provide complex and multi-faceted responses, including \texttt{WatchTime}~ and various types of interactions with multiple videos. On the one hand, the platforms aim at optimizing the users' cumulative \texttt{WatchTime} ~(main goal) in the long term, which can be effectively optimized by Reinforcement Learning. On the other hand, the platforms also need to satisfy the constraint of accommodating the responses of multiple user interactions (auxiliary goals) such as \texttt{Like}, \texttt{Follow}, \texttt{Share}, etc. In this paper, we formulate the problem of short video recommendation as a Constrained Markov Decision Process (CMDP).  We find that traditional constrained reinforcement learning algorithms fail to work well in this setting. We propose a novel two-stage constrained actor-critic method: At stage one, we learn individual policies to optimize each auxiliary signal. In stage two, we learn a policy to (i) optimize the main signal and (ii) stay close to policies learned in the first stage, which effectively guarantees the performance of this main policy on the auxiliaries. Through extensive offline evaluations, we demonstrate the effectiveness of our method over alternatives in both optimizing the main goal as well as balancing the others. We further show the advantage of our method in live experiments of short video recommendations, where it significantly outperforms other baselines in terms of both \texttt{WatchTime}~ and interactions. Our approach has been fully launched in the production system to optimize user experiences on the platform. 
\end{abstract}

\maketitle


\section{Introduction}
\begin{figure}
    \centering
    \includegraphics[width=0.8\linewidth]{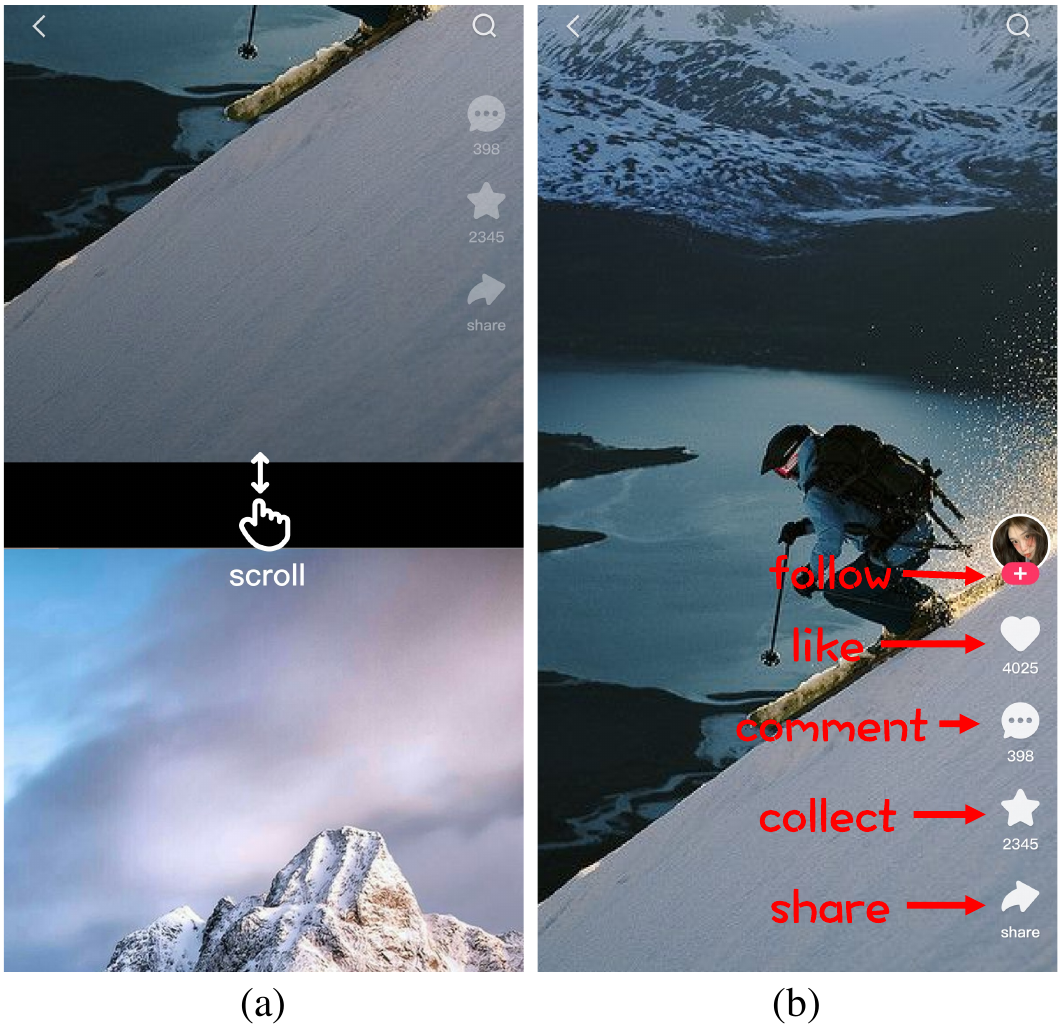}
    \caption{An example of a popular short video (TikTok, Kuaishou, etc) platform.}
    \Description{}
    \label{fig:short_video}
\end{figure}

The surging popularity of short videos has been changing the status quo of social media. Short video consumption has brought in huge business opportunities for organizations. As a result, there has been an increasing interest in optimizing recommendation strategies \cite{wang2022make, zhan2022deconfounding,lin2022feature, gong2022real} for short video platforms. Users interact with the platform by scrolling up and down and watching multiple videos as shown in Figure \ref{fig:short_video}(a). Users provide multi-dimensional responses at each video. As shown in the left part of Figure \ref{fig:short_video}(b), potential responses from a user after consuming a video include \texttt{WatchTime}~(the time spent on watching the video), 
and several types of interactions:
\texttt{Follow}~ (follow the author of the video),
\texttt{Like}~ (Like this video), 
\texttt{Comment}~(provide comments on the video),
\texttt{Collect}~(Collect this video),
\texttt{Share}~(share this video with his/her friends),  etc.

On the one hand, the main goal of the platform is to optimize the cumulative \texttt{WatchTime}~ of multiple videos, as \texttt{WatchTime}~ reflects user attention and is highly related to daily active users (DAU). Recently, a growing literature has focused on applying reinforcement learning (RL) to recommender systems, due to its ability to improve cumulative reward \cite{nemati2016optimal,zhao2017deep,  zhao2018recommendations,chen2018stabilizing, zou2019reinforcement,liu2019deep, chen2019large, xian2019reinforcement, ma2020off, afsar2021reinforcement, ge2021towards, gao2022cirs,wang2022surrogate, xin2022supervised}. In particular, \texttt{WatchTime}~, can be effectively cumulatively maximized to increase user spent time across multiple videos with RL approaches. On the other hand, other responses such as \texttt{Like}/\texttt{Follow}/\texttt{Share}~ also reflect user satisfaction levels. Thus the platform needs to satisfy the constraints of user interactions. Thereby, established recommender systems that exclusively optimize a single objective (such as gross merchandise volume for e-commerce platforms \cite{pi2020search}) is no longer sufficient---the applied systems should take all aspects of responses into consideration to optimize user experiences. 

In this paper, we model the problem of short video recommendation as a Constrained Markov Decision Process: users serve as the environments, and the recommendation algorithm is the agent; at each time step the agent plays an action (recommend a video to the user), the environment sends multiple rewards (responses) to the agent. The objective of the agent is to maximize the cumulative \texttt{WatchTime}~ (main goal) subject to the constraints of other interaction responses (auxiliary goals). Our aim is different from Pareto optimality that aims to find a Pareto optimal solution \cite{sener2018multi,lin2019pareto,chen2021reinforcement}, which may not prioritize the main goal of the system.

The problem of this constrained policy optimization is much more challenging as compared to its unconstrained counterpart. A natural idea would be applying standard constrained reinforcement learning algorithms that maximize the Lagrangian with pre-specified multipliers \cite{tessler2018reward}. However, such method can not apply to our setting for the following two reasons:

First, it is not sufficient to use a single policy evaluation model to estimate 
the Lagrangian dual objective due to different types of responses from the user. 
Such response combination is not adequate, particularly for responses with their own discount factors---the formulation of temporal difference error in value-based models  only allows for a single discount value. In scenarios where one discount factor suffices, it can still be difficult for a single value model to evaluate the policy accurately, especially when different responses are observed at various frequencies, as typical for short video recommendations. The \texttt{WatchTime}~response is dense and observed from each video view, while the interaction-signal such as \texttt{Like}/\texttt{Follow}/\texttt{Share}~is much more sparse and may not be provided within dozens of views. The signal from the sparse responses will be weakened by the dense responses when naively summing them up together. To address this multi-response evaluation difficulty, we separately evaluate each response via its own value model, which allows for response-specific discount factors and mitigates the interference on evaluation from one response on another. Experiments in Section \ref{appendix:offline-critic} validates the effectiveness of this method.

Second, different from only one constraint is considered in \cite{tessler2018reward}, multiple constraints exist in recommender systems, especially in short video systems. We find that it is more difficult for algorithms that maximize the Lagrangian to optimize due to larger search space of multi-dimensional Lagrangian multipliers. It is time costly to grid search on the Lagrangian multipliers as the training of reinforcement learning algorithms takes long time. On account of this, we propose to firstly learn policies to optimize each auxiliary response and then ``softly'' regularize the policy of the main response to be close to others instead of searching optimal value of Lagrangian multipliers. We theoretically prove the closed form of the optimal solution. We demonstrate empirically that our approach can better maximize the main response and balance other responses in both offline and live experiments. 

Together, we summarize our contributions as below:
\begin{itemize}
    \item \textbf{Constrained Optimization in Short Video Recommendations}: We formalize the problem of constrained policy learning in short video recommendations, where different responses may be observed at various frequencies, and the agent maximizes one with the constraint of balancing others.
    \item \textbf{Two-Stage Constrained Actor-Critic Algorithm} We propose a novel two-stage constrained actor-critic algorithm that effectively tackles the challenge: (1) Multi-Critic Policy Estimation: To better evaluate policy on multiple responses that may differ in discount factors and observation frequencies, we propose to separately learn a value model to evaluate each response. (2) Two-Stage Actor Learning: We propose a two-stage actor learning method which firstly learns a policy to optimize each auxiliary response and secondly softly regularizes the policy of the main response to be not far from others, which we demonstrate to be a more effective way in constrained optimization with multiple constraints as compared with other alternatives. 
    \item \textbf{Significant Gains in Offline and Live Experiments}: We demonstrate the effectiveness of our method in both offline and live experiments.
    \item \textbf{Deployment in real world short video application}: We fully launch our method in a popular short video platform.
\end{itemize}

\section{Related Work}

\paragraph{Reinforcement Learning for Recommendation} There is a growing literature in applying RL to recommender systems, for its ability to optimize user long-term satisfaction \cite{afsar2021reinforcement}. Value-based approaches estimate user satisfaction of being recommended an item from the available candidate set and then select the one with the largest predicted satisfaction \cite{nemati2016optimal,zhao2018recommendations,liu2019deep,chen2018stabilizing}. Policy-based methods directly learn the policy (which item to recommend) and optimize it in the direction of increasing user satisfaction \cite{chen2019top,xian2019reinforcement,chen2019large,ma2020off}. Recently, growing attention has been paid to adapting reinforcement learning for more complex recommendation applications beyond optimizing one single objective, such as promoting equal exposure opportunities for content items \cite{ge2021towards}, increasing diversity and novelty of recommendations \cite{stamenkovic2021choosing}, and characterizing more comprehensive user dynamics with  representational reward shaping \cite{chen2021reinforcement}; we view our work as complementary to the third line. 
In face of the multi-faceted user responses, the system in real applications often has preferences on different types of user responses, for which we propose the constrained optimization problem in contrast to pursuing the Pareto optimality as proposed in \cite{chen2021reinforcement} and \cite{ge2022toward}.

\paragraph{Constrained Reinforcement Learning} Our work is also closely related to the literature of  constrained reinforcement learning, where the sequential decision making problem is formulated into a constrained Markov Decision Process \cite{sutton2018reinforcement}, and the policy learning procedure is expected to respect the constraints\cite{liu2021policy,garcia2015comprehensive, chow2017risk,tessler2018reward,dalal2018safe}. As an example, \cite{tessler2018reward} propose to update the policy and the Lagrangian multiplier alternatively and prove the convergence of their algorithm to a fixed point. This approach however only models one constraint, and can not scale well on problems with multiple constraints. In contrast, for each auxiliary response, we learn a policy to maximize it specifically, then we  ``softly'' regularize the main policy to be close to others. We show empirically that this is a more effective way for constrained policy learning when dealing with multiple responses in recommender systems. Different from \cite{nair2020awac} that studies in offline RL and regularizes the learned policy to be near to one behavior policy, we softly restrict the policy within other policies maximizing other auxiliary responses.

\paragraph{Multi-objective Optimization} We also discuss a relevant line on multi-objective optimization. To trade off different objectives, methods in this field can be broadly categorized into two classes: the Pareto optimization and the joint optimization with pre-specified weights. The goal of Pareto optimization is to find a solution such that no other solutions can concurrently improve all objectives, named as \emph{Pareto optimality} \cite{nguyen2020multi,sener2018multi,chen2021reinforcement,ge2022toward}. However, a Pareto optimal solution may not prioritize the objective that is most valued in applications. 
The other method combines different objectives together into a single one via pre-specifying the weights \cite{white1980solution,mossalam2016multi}. However, it is difficult to quantify these weights that can accurately reflect preferences in real applications \cite{tessler2018reward}.

\section{Constrained Markov Decision Process for Short Video Recommendation}

 \begin{figure}[!h]
    \centering
    \includegraphics[width=1.0\linewidth]{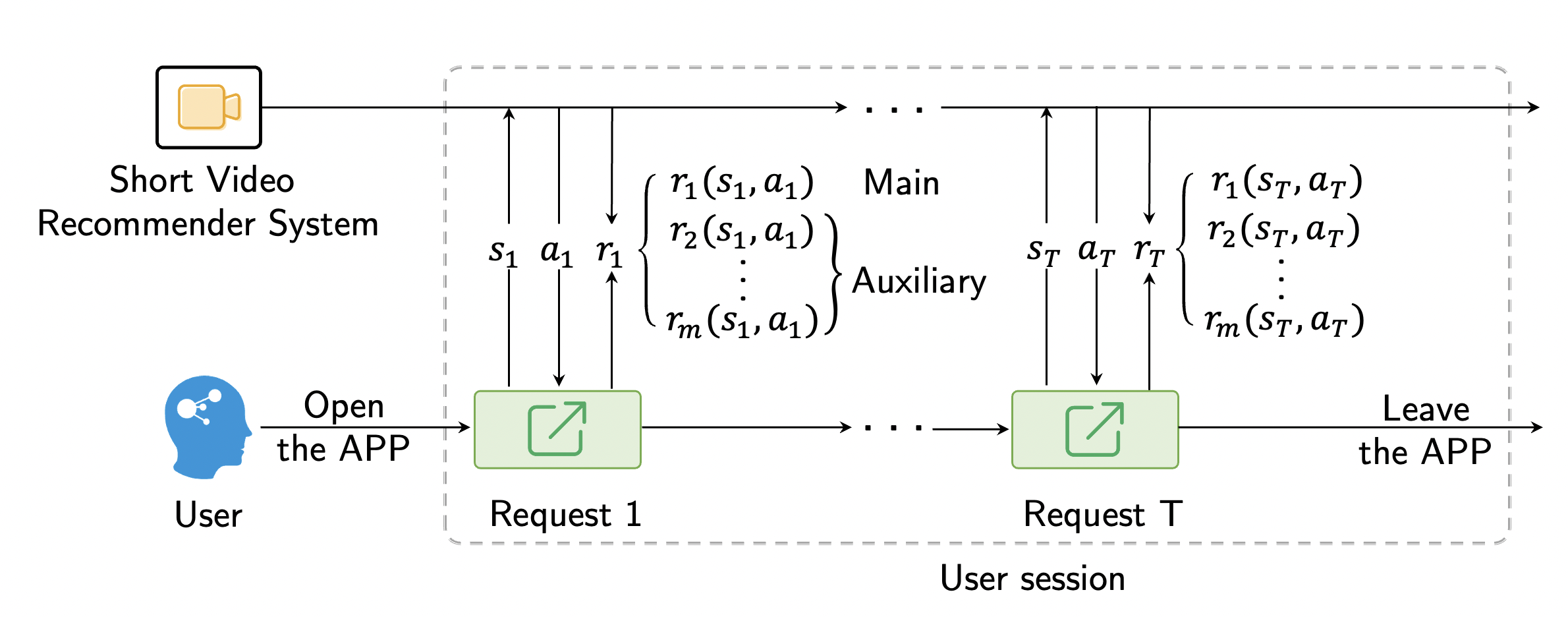}
    \caption{The MDP of short video recommendation.}
    \Description{}
    \label{fig:mdp}
\end{figure}

\label{sec:cmdp}
We start by formulating the problem of short video recommendation, which is shown in
Figure \ref{fig:mdp}. When a user $u$ opens the app, a new \emph{session} starts. A session consists of multiple \emph{requests}. At each request $t$ the recommender system (agent) takes an \emph{action} $a_t$ that recommends the user a video based on the user current \emph{state}. Then the user provides \emph{multi-faceted} responses (such as  \texttt{WatchTime}, \texttt{Like}, \texttt{Share}, and \texttt{Follow}) on the shown video, which are received by the agent as vector-valued \emph{reward} signal. After the user leaves the app, the session ends. The goal of the recommender system is to optimize cumulative reward of the main response (\emph{e.g.}, \texttt{WatchTime}), with the constraint of not sacrificing others much.

We model the above procedure as a Constrained Markov Decision Process (CMDP) \cite{sutton2018reinforcement} $(S, A, P, R, C, \rho_0, \Gamma)$, 
where $S$ is the state space of user current representation $s_t$, 
$A$ is the action space (and each action $a_t$ corresponds to a recommended  video for one request), 
$P:S\times A \rightarrow \Delta(S)$ captures  the state transition, 
$R: S\times A \rightarrow  \mathbb{R}^m$ defines the vector-valued reward function that yields $m$ different rewards $r(s_t, a_t)=\big(r_{1}(s_t, a_t), \dots, r_{m}(s_t, a_t)\big)$, 
$\rho_0$ is the initial state distribution, 
$\Gamma=(\gamma_{1}, \dots, \gamma_{m}) \in (0,1)^m$ denotes the  vector of discount factor for reward of each response. $C$ specifies the constraints on the auxiliary responses, which denotes the lower bound of the total numbers of signals of other objectives.

Define the  vector-valued discounted cumulative reward $R_t$  as ${{R}}_t = \sum_{t'=t}^T \Gamma^{t'-t} \cdot { r}(s_{t'}, a_{t'})$,
where $T$ is the session length (i.e., the number of requests), $\Gamma^{b}=\big(\gamma_{1}^b, \dots, \gamma_{m}^b\big)$, and $\mathbf{x}\cdot \mathbf{y}$ denotes the pointwise product. Let $V^{\pi}(s)=\big(V_{1}^\pi(s),\dots,  V_{m}^\pi(s)\big)$ be the state value $E_{\pi}[R_t|s_t=s]$ under actions sampled in accordance with policy $\pi$ and $Q(s, a)=\big(Q_{1}^\pi(s,a),\dots,  Q_{m}^\pi(s,a)\big)$
be its state-action value $E_{\pi}[R_t|s_t=s, a_t=a]$. Denote $\rho_\pi$ as the state distribution induced by policy $\pi$.
Without loss of generality, we set the first response as our main response. The goal is to learn a recommendation policy $\pi(\cdot|s)$ to solve the following optimization problem: 
\begin{equation}
\label{eq:prob}
\begin{split}
    \max_\pi \quad &  E_{\rho_\pi}\big[V^{\pi}_{1}(s)\big] \\
    \mbox{s.t.}   \quad &  E_{\rho_\pi}\big[V^{\pi}_{i}(s)\big]  \geq C_{i}, \quad i=2,\dots,m
\end{split}
\end{equation}
where $ C_{i}$ is constraint on the \emph{auxiliary} response $i$.

\section{Two-Stage Constrained Actor-Critic}
In this section, we propose a novel two-stage constrained actor-critic method, addressing the learning challenges in the context of short video recommendation:
\begin{description}
    \item[Multi-Critic Policy Estimation] We propose to estimate the responses separately to better estimate dense and sparse signals. 
    \item[Stage One] For each auxiliary response, we learn a policy to optimize its cumulative reward.  
    \item[Stage Two] For the main response, we learn a policy to  optimize its cumulative reward, while softly limiting it to be close to other policies that are learned to optimize the auxiliary. 
\end{description}
We first discuss the advantage of evaluating different policies separately over estimating jointly. Secondly, we elaborate our method in the settings of online learning with stochastic policies in Sections \ref{sec:stage-one} and \ref{sec:stage-two}. We then discuss its extensions to the offline setting and deterministic policies.

\subsection{Multi-Critic Policy Estimation}
\label{appendix:offline-critic}
We showcase the advantage of separate evaluation for each response over a joint evaluation of summed response. Specifically, we consider two types of responses from each video view: \texttt{WatchTime}~and interactions (which is an indicator function of whether the interactions happen during the view).
\begin{itemize}
    \item For the joint evaluation, we learn a value model $V_{joint}$ with reward as a sum of \texttt{WatchTime}~and interactions.
    \item For the separate evaluation, we learn two value models  $V_w$ and $V_i$ with reward as \texttt{WatchTime}~and interactions respectively. Define the value  of separate evaluation as $V_{separate}=V_w + V_i$~
\end{itemize}
For fair comparison, we share the same discount factor $0.95$ for all value models and train them on the same data collected from a popular short video platform for one day. To evaluate the accuracy of the value model in terms of \texttt{WatchTime}~and interactions, we compute the correlation between model values $V_{joint}$ and $V_{separate}$ with the Monte Carlo value of the sum of the corresponding responses in each session. As compared to $V_{joint}$, $V_{separate}$ is more correlated with \texttt{WatchTime}~ and interactions by $0.19\%$ and $0.14\%$ respectively(a $0.1\%$ improvement on \texttt{WatchTime}~ and interactions is significant), demonstrating that the separate evaluation better learns different reward responses than jointly learning.

\subsection{Stage One: Policy Learning for Auxiliary Responses}
\label{sec:stage-one}
At this stage, we learn policies to optimize the cumulative reward of each auxiliary response separately. For completeness, we write out our procedure for stochastic policies \cite{williams1992simple}. Considering response $i$, let the learned actor and the critic be parameterized by $\pi_{\theta_i}$ and $V_{\phi_i}$ respectively. At iteration $k$, we observe sample $(s,a,s')$ collected by $\pi_{\theta_i^{(k)}}$, \emph{i.e.}, $s\sim \rho_{\pi_{\theta_i^{(k)}}}, a\sim \pi_{\theta_i^{(k)}}(\cdot|s)$ and $s'\sim P(\cdot|s,a)$. We update the critic to minimize the Bellman equation:
\begin{equation}
\label{eq:nonmajor-critic}
    \phi_{i}^{(k+1)} \leftarrow \arg\min_{\phi}
    E_{\pi_{\theta_i^{(k)}}}\Big[
    \big(r_{i}(s, a) + \gamma_{i}V_{\phi_i^{(k)}}(s') -V_{\phi}(s)  \big)^2
    \Big].
\end{equation}
We update the actor to maximize the advantage:
\begin{equation}
\label{eq:nonmajor-actor}
\begin{split}
    &\theta_{i}^{(k+1)} \leftarrow \arg\max_{\theta}  E_{\pi_{\theta_i^{(k)}}}\Big[ A_{i}^{(k)}\log\big(\pi_\theta(a|s)\big)\Big]\\
    \mbox{where}\quad & A_{i}^{(k)} = r_{i}(s, a) +\gamma_{i} V_{\phi_i^{(k)}}(s') -V_{\phi_i^{(k)}}(s).
\end{split}
\end{equation}

\subsection{Stage Two: Softly Constrained Optimization of the Main Response}
\label{sec:stage-two}
After pre-training the policies $\pi_{\theta_2}, \dots, \pi_{\theta_m}$ that optimize the auxiliary responses, we now move onto the second stage of learning the policy to optimize the main response. We propose a new constrained policy optimization method with multiple constraints.

Let the actor and the critic be $\pi_{\theta_1}$ and $V_{\phi_1}$ respectively. At iteration $k$, we similarly update the critic to minimize the Bellman equation:
\begin{equation}
\label{eq:major-critic}
     \phi_{1}^{(k+1)} \leftarrow \arg\min_{\phi}
    E_{\pi_{\theta_1^{(k)}}}\Big[
    \big(r_{1}(s, a) + \gamma_{1}V_{\phi_1^{(k)}}(s') -V_\phi(s)  \big)^2
    \Big].
\end{equation}

The principle of updating the actor is two-fold: (i) maximizing the advantage; (ii) restricting the policy to the domain that is not far from other policies. The optimization is formalized below:
\begin{equation}
\label{eq:awac}
    \begin{split}
         \max_{\pi}\quad  & E_{\pi}[ A_{1}^{(k)}]\\
         \mbox{s.t.} \quad & D_{KL}(\pi|| \pi_{\theta_i})\leq \epsilon_i,\quad i=2,\dots,m,\\
    \mbox{where}\quad & A_{1}^{(k)} = r_{1}(s, a) +\gamma_{1} V_{\phi_1^{(k)}}(s') - V_{\phi_1^{(k)}}(s).
    \end{split}
\end{equation}

We get the closed form solution of the Lagrangian of Eq. (\ref{eq:awac}) in the following theorem. We omit the proof due to lack of space, please refer to Appendix \ref{proof}.

\begin{theorem}
The Lagrangian of Eq. (\ref{eq:awac}) has the closed form solution 
\begin{equation}
  \pi^*(a|s)  \propto \prod_{i=2}^m \big(\pi_{\theta_i}(a|s)\big)^{\frac{\lambda_i}{\sum_{j=2}^m \lambda_j}}\exp\bigg(\frac{A_{1}^{(k)}}{\sum_{j=2}^m \lambda_j} \bigg),
\end{equation}
where $\lambda_i$ with $i=2,\dots,m$ are Lagrangian multipliers.
\end{theorem}

Given data collected by $\pi_{\theta_1^{(k)}}$, we learn the policy $\pi_{\theta_1}$ by minimizing its KL divergence from the optimal policy $\pi^*$:
\begin{equation}
\label{eq:major-actor}
\begin{split}
     &\theta_1^{(k+1)}\leftarrow \arg\min_\theta E_{\pi_{\theta_1^{(k)}}}[D_{KL}(\pi^*(a|s)||\pi_\theta(a|s))]\\
     = &\arg\max_\theta E_{\pi_{\theta_1^{(k)}}}\Big[ 
      \frac{\prod_{i=2}^m\Big({\pi_{\theta_i}(a|s)}\Big)^{\frac{\lambda_i}{\sum_{j=2}^m \lambda_j}}}{{\pi_{\theta_1^{(k)}}(a|s)}}
      \exp\bigg(\frac{A_{1}^{(k)}}{\sum_{j=2}^m \lambda_j} \bigg)
    \log \pi_\theta(a|s) \Big].
\end{split}
\end{equation}

The procedure of the two-stage constrained actor-critic algorithm is shown in Appendix \ref{alg}, and we name it as TSCAC for short. We here provide some intuition behind actor updating in \eqref{eq:major-actor}. The term ${\pi_{\theta_i}(a|s)}$ denotes 
the probability the action selected by policy $i$ and serves as an importance,
which softly regularizes the learned policy $\pi_{\theta_1}$ to be close to other policies $\pi_{\theta_i}$. Smaller Lagrangian multipliers $\lambda_i$ indicate weaker constraints, and when $\lambda_i=0$, we allow the learned policy $\pi_{\theta_1}$ to be irrelevant of the constraint policy $\pi_{\theta_i}$. 
Note that we set the value of $\lambda$ to be the same, which is more practical for the production system. The performance of TSCAC would be better if we fine-tune it with different Lagrangian multiplier value. But the effectiveness of TSCAC with the same value of $\lambda$ is validated in both offline and live experiments, as we will see in following sections.

\textbf{Offline Learning}
We now discuss adapting our constrained actor-critic method to the offline setting, i.e., a fixed dataset. The main change when moving from the online learning to the  offline learning is the bias correction on the policy gradient. The actor is no longer updated on data collected by current policy but by another behavior policy $\pi_\beta$, which may result in a different data distribution induced by the policy being updated.  To address the distribution mismatch when estimating the policy gradient, a common strategy is to apply bias-correction ratio via importance sampling \cite{precup2000eligibility,precup2001off}. Given a trajectory $\tau=(s_1,a_1, s_2, a_2, \dots)$, the bias-correction ratio on the policy gradient for policy $\pi_{\theta_i}$ is
$
\label{eq:unbiased-is}
    w(s_t, a_t) = \prod_{t'=1}^t \frac{\pi_{\theta_i}(s_{t'}|a_{t'})}{\pi_\beta(s_{t'}|a_{t'})},
$
which gives an unbiased estimation, but the variance can be huge. Therefore, we suggest using a first-order approximation, and using the current action-selection ratio when optimizing the actors of auxiliary responses, 


\begin{equation}
    \begin{split}
        \theta_{i}^{(k+1)} \leftarrow&   \arg\max_{\theta}  E_{\pi_\beta}\bigg[\frac{\pi_{\theta_i^{(k)}}(a|s)}{ \pi_\beta(a|s)} A_{i}^{(k)}\log(\pi_\theta(a|s))\bigg].
    \end{split}
\end{equation}


When updating the actor of the main response, we have
\begin{equation}
    \begin{split}
         \theta_{1}^{(k+1)}  \leftarrow  
     &\arg\max_\theta E_{ \pi_\beta}\Big[\frac{
     \prod_{i=2}^m \Big(\pi_{\theta_i}(a|s)\Big)^{\frac{\lambda_i}{\sum_{j=2}^m \lambda_j}}}{\pi_{\beta}(a|s)}\\
     &\qquad\qquad\times
      \exp\bigg(\frac{A_{1}^{(k)}}{\sum_{j=2}^m \lambda_j} \bigg)
    \log( \pi_\theta(a|s)) \Big].
    \end{split}
\end{equation}

\textbf{Deterministic Policies}
\label{sec:ddpg}
We now discuss the extension of TSCAC to deterministic policies\cite{lillicrap2015continuous}, inspired by the updating rule for the actor of constrained policy discussed in \eqref{eq:major-actor}. 
Similarly, at stage one, for each auxiliary response $i$, we learn separate 
critic models $Q_{\phi_i}(s,a)$ and actor models $\pi_{\theta_i}(s)$.  At stage two, for the main response, we learn critic $Q_{\phi_1}(s,a)$ via temporal learning, and for actor $\pi_{\theta_1}(s)$, the updating rule follows the form: 

\begin{equation}
    \max_{\theta}\quad \prod_{i=2}^m \bigg( h(\pi_{\theta_i}(s),\pi_{\theta_1}(s))\bigg)^{\lambda_i} Q_{\phi_1}(s, \pi_{\theta}(s)),
    \label{eq:ddpg}
\end{equation}
where $h(a_1, a_2)$ scores high when two actions $a_1, a_2$ are close to each other and scores low vice versa, and $h(\pi_{\theta_i}(s),\pi_{\theta_1}(s))$ scores high when the actions selected by policy $\pi_{\theta_1}$ and $\pi_{\theta_i}$ are close. $\lambda_i\geq 0$ plays a similar role as the constraint Lagrangian multiplier---larger $\lambda_i$ denotes stronger constraint. As an example, given $n$ dimensional action space, one can choose $h(a_1, a_2)=\sum_{d=1}^{n}\exp\big(-\frac{(a_{1d}-a_{2d})^2}{2}\big)$.
The deterministic version of TSCAC can apply to the setting with continuous actions, such as the embedding of the user preference. 

\section{Offline Experiments}
\label{sec:offline}
In this section, we evaluate our method on a public dataset about short video recommendation via extensive offline learning simulations. We demonstrate the effectiveness of our approach as compared to existing baselines in both achieving the main goal and balancing the auxiliaries. We also test the versatility of our method on another public recommendation dataset, please refer to Appendix \ref{sec:offline} due to lack of space. 

\subsection{Setup}
\paragraph{Dataset} We consider a public dataset for short video recommendation named \emph{KuaiRand} (\url{https://kuairand.com/})~\citep{gao2022kuairand}, which is collected from a famous video-sharing mobile app and suitable for the offline evaluation of RL methods as it is unbiased. This dataset collects not only the overall \texttt{WatchTime}~ of the videos, but also the interaction behavior of the users including \texttt{Click}, \texttt{Like}, \texttt{Comment}~ and \texttt{Hate}. The statistics of the dataset are illustrated in Table \ref{tab: dataset}. It shows that \texttt{Like}, \texttt{Comment}, and \texttt{Hate}~ are sparse signals. Note that \texttt{Hate}~ is extremely sparse. Logs provided by the same user are concatenated to form a trajectory; we choose top $150$ videos that are most frequently viewed. 


\begin{table}[t]
    \centering
    \caption{The statistics of KuaiRand.}
    \begin{tabular}{c|c|c}
        \hline
        Dimension & Number & Sparse Ratio \\
        \hline
        users& 26858& -
        \\
        items& 10,221,515& -
        \\
        samples& 68,148,288& -
        \\
        click & 25,693,008& 37.70\%
        \\
        like &1094434& 1.61\%
        \\
        comment & 163977& 0.24\%
        \\
        hate &32449& 0.048\%
        \\
        \hline
    \end{tabular}
    \label{tab: dataset}
\end{table}

\paragraph{MDP} 
\begin{itemize}
    \item state $s_t$: A 1044 dimension vector, which is a concatenation of user features(user property),  the last $20$ video features viewed by the user(user history) and all the $150$ candidate video features(context). 
    \item action $a_t$: the video ID to be recommended currently.
    \item reward $r_t$: a vector of five scores the user provided for the viewed videos in terms of \texttt{Click}, \texttt{Like}, \texttt{Comment}, \texttt{Hate}, and \texttt{WatchTime}. 
    \item episode: a sequence of users' video viewing history.
    \item discount factor $\gamma$: 0.99
    \item objective: We set the main goal to be maximizing the video \texttt{WatchTime}, and treat others as the auxiliaries. 
\end{itemize}

\paragraph{Evaluation}  We use the \emph{Normalised Capped Importance Sampling} (NCIS) approach to evaluate different policies, which is a standard offline evaluation approach for RL methods in recommender systems \cite{zou2019reinforcement}. We also evaluate our method in terms of other metrics, please refer to Appendix \ref{sec: other metrics}. The NCIS score is defined:
\begin{equation}
\label{ncis}
N(\pi)=\frac{\sum_{s,a\in D}w(s,a)r(s,a)}{\sum_{s,a\in D}w(s,a)}, w(s,a)=\min
\{c,\frac{\pi(a|s)}{\pi_{\beta}(a|s)}\},
\end{equation}
where $D$ is the dataset, $w(s,a)$ is the clipped importance sampling ratio, $\pi_{\beta}$ denotes the behavior policy, $c$ is a positive constant.


\paragraph{Baselines}\label{sec:alg} We compare TSCAC with the following baselines.
\begin{itemize}
    \item \textbf{BC}: A supervised behavior-cloning policy $\pi_\beta$ to mimic the recommendation policy in the dataset, which inputs the user state and outputs the video ID. 
    
    \item \textbf{Wide\&Deep} \cite{cheng2016wide}: A supervised model which utilizes wide and deep layers to balance both memorization and generalization, which inputs the user state, outputs the item id, and the weight of each sample is set to be the weighted sum of all responses of this item. 
    
    \item \textbf{DeepFM} \cite{guo2017deepfm}: a supervised recommendation model which combines deep neural network and factorization machine, which inputs the user state, outputs the item id, and the weight of each sample is set to be the weighted sum of all responses of this item.
    
    \item \textbf{RCPO} \cite{tessler2018reward} : A constrained actor-critic approach called reward-constrained policy optimization which optimizes the policy to maximize the Lagrange dual function of the constrained program. Specifically, the reward function is defined as $r=r_0+\sum_{i=1}^{n}\lambda_i*r_i$, where $r_0$ is main objective, \texttt{WatchTime} and $r_i$ denotes other feedback, and $\lambda_i$ is the Lagrangian Multiplier.

    \item \textbf{RCPO-Multi-Critic}: We test an improved version of RCPO with multiple critics. We separately learn multiple critic models to evaluate the cumulative rewards of each feedback. Then when optimizing the actor, we maximize a linear combination of critics, weighted by the Lagrangian multipliers.

    \item \textbf{Pareto} \cite{chen2021reinforcement}: A  multi-objective RL algorithm that finds the Pareto optimal solution for recommender systems.
    \item \textbf{TSCAC}: our two-stage constrained actor-critic algorithm. 
\end{itemize}


\subsection{Overall Performance}

\begin{table*}[htb]
    \centering
    \caption{Performance of different algorithms on KuaiRand.}
    \begin{tabular}{c|c|c|c|c|c}
    \toprule
        Algorithm  &  Click$\uparrow$ & Like$\uparrow$(e-2) &  Comment$\uparrow$(e-3)
        & Hate$\downarrow$(e-4) & WatchTime$\uparrow$  \\
    \midrule
    BC &$0.5338$ &$1.231$ &$3.225$ &$2.304$ &$12.85$
    \\
    \hline
     Wide\&Deep & $0.5544$ &$1.244$& $3.344$ & $2.011$ & $12.84$
     \\
     &$3.86\%$ & $1.07\%$& $3.69\%$& $-12.7\%$& $-0.08\%$
     \\
      \hline
    DeepFM  &$0.5549^{*}$ &$1.388^{*}$ &$3.310$ &$2.112$ & $12.92$
    \\
    &$3.95\%^{*}$ & $12.76\%^{*}$&  $2.64\%$& $-8.31\%$ & $0.53\%$ 
    \\
    \hline 
    RCPO &  $0.5510$ & $1.386$ & $3.628^{*}$ & $2.951$ & $13.07^{*}$
    \\
    &$3.23\%$ & $12.57\%$ & $12.5\%^{*}$ & $28.1\%$ & $1.70\%^{*}$
    \\
     \hline 
    RCPO-Multi-Critic & $0.5519$ & $1.367$ & $3.413$& $2.108$& $13.00$
    \\
    & $3.41\%$ & $11.04\%$ & $5.83\%$& $-8.49\%$& $1.14\%$
    \\ 
    \hline 
    Pareto & $0.5438$ & $1.171$ & $3.393$ & $0.9915^{*}$ & $11.90$
    \\
    & $1.87\%$ & $-4.85\%$ & $5.22\%$ & $-56.96\%^{*}$ & $-7.4\%$
    \\ 
    \hline 
    TSCAC & ${\bf{0.5570}}$& ${\bf{1.462}}$ & ${\bf{3.728}}$& $1.870$& 
    ${\bf{13.14}}$
    \\
    &{\bf{4.35\%}}& ${\bf{18.80\%}}$& ${\bf{15.6\%}}$& $-18.83\%$& 
    ${\bf{2.23\%}}$
    \\ 
    \bottomrule
    \end{tabular}
    \\The number in the bracket stands for the unit of this column; The number in the first row of each algorithm is the NCIS score.
\\The percentage in the second row  means the performance gap between the algorithm and the BC algorithm.
\\The numbers with $*$ denote the best performance among all baseline methods in each response dimension.
\\The last row is marked by bold font when TSCAC achieves the best performance at each response dimension.
\label{tab:offlineV2}
\end{table*}

Table \ref{tab:offlineV2} presents the performance of different algorithms in terms of five scores. We can see that our TSCAC algorithm significantly outperforms other algorithms including both constrained reinforcement learning and supervised learning methods: for the main goal (\texttt{WatchTime}), TSCAC achieves the highest performance $13.14$($2.23\%$); for the auxiliary goal,  TSCAC also ranks highest for $3$ out of $4$ scores (\texttt{Click}, \texttt{Like}, \texttt{Comment}). Note that TSCAC outperforms BC and RCPO at each dimension. The Pareto algorithm indeed learns a Pareto optimal solution that achieves best performance at \texttt{Hate}, but gets the lowest performance $11.90$($-7.4\%$), i.e., it does not satisfy the setting with the main goal to optimize the \texttt{WatchTime}. The RCPO algorithm achieves the second highest performance at \texttt{WatchTime}, $13.07(1.70\%)$, but the score at \texttt{Hate}~ is the worst as the sparse signals are dominated by dense signals in a single evaluation model. Compared with RCPO, RCPO-Multi-Critic achieves much better score at \texttt{Hate}, which demonstrates the effectiveness of the multi-critic policy estimation method. TSCAC also outperforms RCPO-Multi-Critic at each dimension, which shows that the ability of our two-stage actor learning method to deal with multiple responses.

\subsection{Ablation Study}
\begin{figure*}
    \centering
    \includegraphics[width=1.\linewidth]{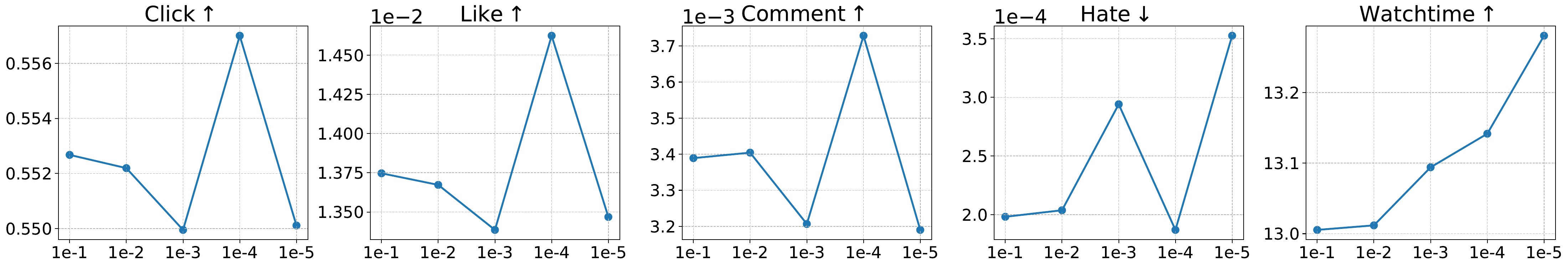}
    \caption{Effect of the value of the Lagrangian multiplier on the performance.}
    \Description{}
    \label{fig:offline-ablation}
\end{figure*}

We investigate how the value of Lagrangian multiplier affects the performance. As we set the value of $\lambda$ of all constraints to be the same in the second stage, we vary $\lambda$ across $[1e-1, 1e-2, 1e-3,1e-4,1e-5]$ and present performance of TSCAC in terms of all responses. Recall that larger $\lambda$ denotes stronger constraints of auxiliary responses. Figure \ref{fig:offline-ablation} shows that with $\lambda$ increasing, the main goal, \texttt{WatchTime}~ decreases as the constraints of auxiliary responses become stronger. As shown in Figure \ref{fig:offline-ablation}, the performance of interactions drops with small $\lambda$ $1e-5$ as the constraints are weak. Interestingly, the performance of interactions also decreases with larger $\lambda$, which shows that too strong constraints affect the learning of the policy. The value of $1e-4$ achieves the best performance at interactions, and improve \texttt{WatchTime}~ significantly compared with other baselines.


\section{Live Experiments}
To demonstrate the effectiveness of our algorithm, we test its performance as well as other alternatives via live experiments in a popular short video platform. Algorithms are embodied in a candidate-ranking system used in production at a popular short video platform, that is, when a user arrives, these algorithms are expected to rank the candidate videos, and the system will recommend the top video to the user. 
We show that the proposed TSCAC algorithm is able to learn a policy that maximizes the main goal while also effectively balancing the auxiliary goal, and in particular, we set the main one as maximizing the \texttt{WatchTime} and the auxiliary one as improving the interactions between users and videos.

\subsection{Setup}
\paragraph{Evaluation metrics}
We use online metrics to evaluate policy performance. For the main goal, we look at the total amount of time user spend on the videos, referred to as \texttt{WatchTime}. For the auxiliary goal, users can interact with videos through multiple ways, such as sharing the video to friends, downloading it, or providing comments. Here, we focus on the three online metrics associated with the user-video interactions---the total number of \texttt{Share}, \texttt{Download}, \texttt{Comment}~interactions.

\begin{figure*}
    \centering
\includegraphics[width=1.\linewidth]{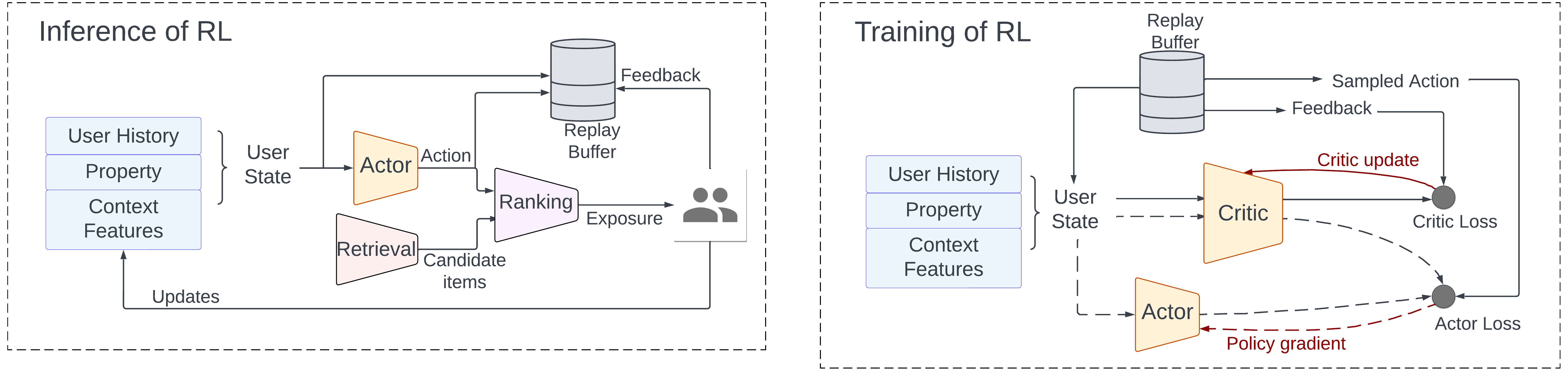}
    \caption{The workflow of RL in production system.}
    \Description{}
    \label{fig:workflow}
\end{figure*}

\paragraph{MDP}
Following the formulation in Section \ref{sec:cmdp}, we present the details of the Constrained MDP for short video recommendation. 
\begin{itemize}
    \item state $s_t$: user historical interactions (the list of items recommended to users at previous rounds and corresponding user feedbacks), user property (such as device and location) and the feature (the embeddings and statistics) of candidate videos at time $t$. 
    \item action $a_t$: a vector embedding of algorithm-predicted user preferences on different video topics, which determines the actual recommendation action(the video to be recommended) via a ranking function described below:

    {\bf the ranking function}: for each candidate video, this function calculates the dot product between the predicted user preference vector ($a_t$) and the video embedding (representing its topic and quality) as in \cite{dulac2015deep}. Then the video with the largest score is recommended.
    \item reward $r_t=(l_t, i_t)$: after each recommendation, the system observes how long the user spent on the video, \texttt{WatchTime}~, denoted as $l_t$, and whether the user has interacted with the video (\texttt{Share}/\texttt{Download}/\texttt{Comment}), denoted as $i_t$.
    \item episode: a trajectory starts when a user opens the app and ends when the user leaves. 
    \item policy: we choose to learn a Gaussian policy in the live experiments. Specifically, the action $a_t$ is sampled from a multivariate Gaussian distribution whose mean and variance are output of the actor model.
\end{itemize}

\paragraph{Workflow}
As shown in Figure \ref{fig:workflow}, RL runs as follows:
\begin{itemize}
\item \textbf{Inference} When the user comes, the user state are sent to the actor network, the actor network sample action by the Gaussian distribution. Then the ranking function inputs both the action and the embedding of candidates, calculates the dot product between the action and the video embeddings as scores, and output the item with the highest score to the user. After that, (state, action, rewards, next state) are saved in the replay buffer.

\item \textbf{Training} The actor and the critic networks are trained with a mini-batch (state, action, rewards, next state), sampled from the replay buffer. 
\end{itemize}

\paragraph{Compared algorithms} 
We complement our evaluation with a supervised learning-to-rank (LTR) baseline, which is the default model run on the platform. 
\begin{itemize}
    \item \textbf{RCPO}:  Following \cite{tessler2018reward}, we define a combined reward $l_t + \lambda i_t$ and learn a policy to maximize the cumulative combined reward with discount factor $0.95$, where $\lambda$ is the Lagrangian multiplier.
    
    \item \textbf{TSCAC}: We first learn a policy $\pi_2$ to optimize the auxiliary goal. 
    Then we learn a policy $\pi_1$ to optimize the main goal with the soft constraint that $\pi_1$ is close to $\pi_2$.
        \begin{itemize}
            \item \textbf{Interaction-AC}: At the first stage, we  learn a policy $\pi_2$ to maximize the interaction reward, with critic update following \eqref{eq:nonmajor-critic} and actor update following \eqref{eq:nonmajor-actor}.
            \item \textbf{TSCAC} At the second stage, we learn a main policy $\pi_1$ to maximize the cumulative reward of \texttt{WatchTime}~ and softly regularize $\pi_1$
            to be close to $\pi_2$, with critic update following \eqref{eq:major-critic} and actor update following \eqref{eq:major-actor}. 
        \end{itemize}
        
    \item \textbf{LTR (Baseline)}: The learning-to-rank model \cite{liu2009learning} that takes user state embedding and video embedding as input and fits the sum of responses.
\end{itemize}


\paragraph{Experimental details} To test different algorithms, we randomly split users on the platform into several buckets. The first bucket runs the baseline LTR model, and the remaining buckets run models RCPO, Interaction-AC, and TSCAC. Models are trained for a couple of days and then are fixed to test performance within one day.

\begin{table}[]
\label{table:result}
\caption{Performance comparison of different algorithms with the LTR baseline in live experiments.}
    \centering
    \begin{tabular}{p{2.3cm}|c|c|c|c}
    \toprule
        Algorithm &  \texttt{WatchTime} & \texttt{Share} &   \texttt{Download} & \texttt{Comment} \\
        \midrule
        RCPO & $+0.309\%$ & $ -0.707\%$ &  $0.153\%$ & $-1.313\%$ \\
        Interaction-AC & $+0.117\%$ & $+5.008\%$ & $+1.952\%$ & $-0.101\%$ \\
        TSCAC & $+0.379\%$ & $+3.376\%$ & $+1.733\%$ & $-0.619\%$ \\
         \bottomrule
    \end{tabular}
    
    \label{tab:live}
\end{table}

\subsection{Results}


Table \ref{tab:live} shows the performance improvement of algorithm comparison with the LTR baseline
regarding metrics \texttt{WatchTime}, \texttt{Share}, \texttt{Download}, and \texttt{Comment}. As we can see, RCPO can learn to improve the \texttt{WatchTime}~ as compared to the baseline; but interaction-signals are too sparse with respect to \texttt{WatchTime}, such that when combining these responses together, it cannot effectively balance the interaction well. Performance of the Interaction-AC algorithm is as expected: with signal from only the interaction reward, it learns to improve the interaction-related metrics (\texttt{Share}, \texttt{Download}, \texttt{Comment}); such interactions between users and videos also improve the user \texttt{WatchTime}, since more interesting videos with high potential of invoking interactions are recommended, which optimizes user whole experience. Finally, The TSCAC algorithm achieves the best performance: as compared to RCPO, it has better \texttt{WatchTime} and does much better on interaction metrics, thanks to the effective softly regularization during training that it should not be too far from the Interaction-AC policy. Note that 0.1\% improvement of \texttt{WatchTime}~ and 1\% improvement of interactions are statistically significant in the short video platform. That is, the performance improvement of our proposed method over baselines is significant. The universal drop of Comment for all RL methods is due to the natural trade-off between \texttt{WatchTime}~ and \texttt{Comment}.

 \begin{figure}
    \centering
    \includegraphics[width=1.\linewidth]{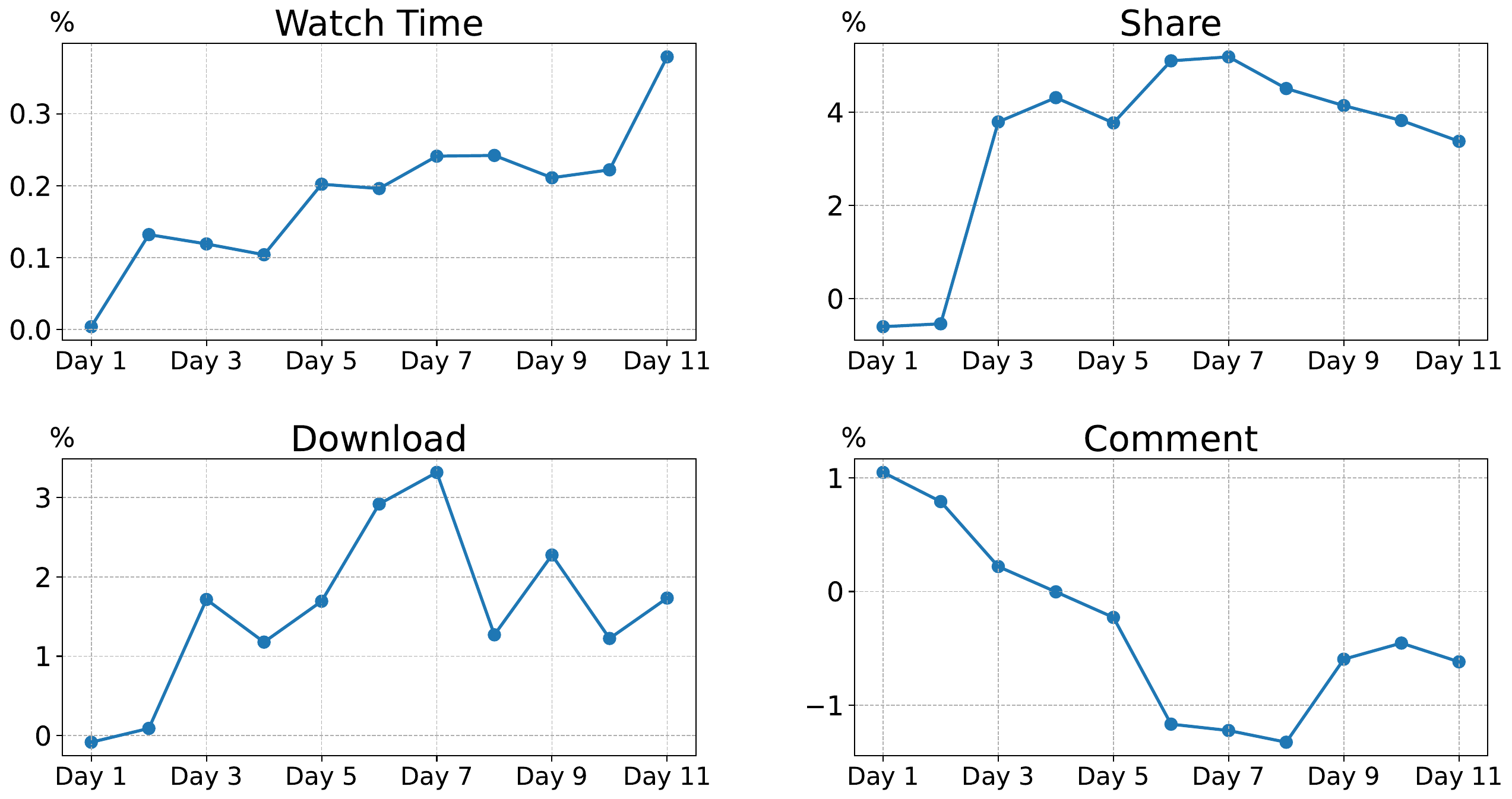}
    \caption{Online performance gap of TSCAC over the LTR baseline of each day. }
    \Description{}
    \label{fig:live_learning}
\end{figure}

To understand how the TSCAC algorithm learns to balance the main and auxiliary goal, Figure \ref{fig:live_learning}  plots the online performance gap of the second stage over the LTR baseline on both \texttt{WatchTime}~ and interactions. As shown, the algorithm quickly learns to improve the interaction metrics \texttt{Share}~ and \texttt{Comment}~ at the beginning, with the constraint of Interaction-AC policy. Then gradually, the model learns to improve \texttt{WatchTime}~over time with sacrificing interactions a little. Note that the live performance of TSCAC outperforms RCPO significantly at each dimension, which demonstrates the effectiveness of our method. 

\section{Conclusion}

In this paper we study the problem to optimize main cumulative responses with multiple auxiliary sparse constraints in short video platforms. To tackle the challenge of multiple constraints, we propose a novel constrained reinforcement learning method, called TSCAC, 
that optimizes the main goal as well as balancing the others for short video platforms. Our method consists of multiple critic estimation and two learning stages. At stage one, for each auxiliary response, we learn a policy to optimize its cumulative reward respectively. At stage two, we learn the major policy to optimize the cumulative main response, with a soft constraint that restricts the policy to be close to policies maximized for other responses. We demonstrate the advantages of our method over existing alternatives via extensive offline evaluations as well as live experiments.  For the future work, it is promising to apply our method to other recommender systems. It is also an interesting future work to study the performance of the deterministic version of TSCAC.


\newpage
\bibliographystyle{ACM-Reference-Format}
\bibliography{theweb2023cameraready}

\newpage
\appendix

\section{Proof of Theorem 1}
\label{proof}
\begin{theorem}
The Lagrangian of Eq. (\ref{eq:awac}) has the closed form solution 
\begin{equation}
  \pi^*(a|s)  \propto \prod_{i=2}^m \big(\pi_{\theta_i}(a|s)\big)^{\frac{\lambda_i}{\sum_{j=2}^m \lambda_j}}\exp\bigg(\frac{A_{1}^{(k)}}{\sum_{j=2}^m \lambda_j} \bigg),
\end{equation}
where $\lambda_i$ with $i=2,\dots,m$ are Lagrangian multipliers.
\end{theorem}


\begin{proof}
The Lagrangian of Eq. (\ref{eq:awac}) is 
\begin{equation}
\label{eq:awac_lagrangian}
     \mathcal{L}(\pi, \lambda_2,\dots, \lambda_m) = -E_{\pi}[ A_{1}^{(k)}] + \sum_{i=2}^m\lambda_i\big(
    D_{KL}(\pi|| \pi_{\theta_i}) - \epsilon_i
    \big).
\end{equation}

Compute the gradient of $ \mathcal{L}(\pi, \lambda_2,\dots, \lambda_m)$ with respect to $\pi$,
\begin{equation}
    \frac{\partial \mathcal{L}}{\partial \pi} = -A_1^{(k)} + \sum_{i=2}^m \lambda_i\big( 1 + \log(\pi(a|s)) - \log(\pi_{\theta_i}(a|s)) \big).
\end{equation}
Setting $ \frac{\partial \mathcal{L}}{\partial \pi}=0$, we have the solution $\pi^*$ satisfies
\begin{equation}
    \pi^*(a|s) = \frac{1}{Z(s)} \prod_{i=2}^m \big(\pi_{\theta_i}(a|s)\big)^{\frac{\lambda_i}{\sum_{j=2}^m \lambda_j}}\exp\bigg(\frac{A_{1}^{(k)}}{\sum_{j=2}^m \lambda_j} \bigg), 
\end{equation}
where $Z(s)$ is the partition function to such that $\int_a \pi^*(a|s)=1$.
\end{proof}

\section{The Two-Stage Constrained Actor-Critic Algorithm}
\label{alg}
\begin{algorithm}
\caption{Two-Stage Constrained Actor-Critic (TSCAC)}
\textbf{Stage One:} For each auxiliary response $i=2,\dots,m$, learn a policy to optimize the response $i$, with  $\pi_{\theta_i}$ denoting actor and $V_{\phi_i}$ for critic.

While not converged, at iteration $k$:
\begin{small}
\begin{equation*}
    \begin{split}
       \phi_{i}^{(k+1)} \leftarrow  & \arg\min_{\phi}
    E_{\pi_{\theta_i^{(k)}}}\Big[
    \big(r_{i}(s, a) + \gamma_{i}V_{\phi_i^{(k)}}(s') -V_{\phi}(s)  \big)^2
    \Big],\\
        \theta_{i}^{(k+1)} \leftarrow & \arg\max_{\theta}  E_{\pi_{\theta_i^{(k)}}}\Big[ A_{i}^{(k)}\log\big(\pi_\theta(a|s)\big)\Big].
    \end{split}
\end{equation*}
\end{small}

\textbf{Stage Two:} For the main response, learn a policy  to both optimize the main response  and restrict its domain close to the policies $\{\pi_{\theta_i}\}_{i=2}^m$ of auxiliary responses, with $\pi_{\theta_1}$ denoting actor and $V_{\phi_1}$ for critic.

While not converged, at iteration $k$:
\begin{small}
\begin{equation*}
    \begin{split}
    \phi_{1}^{(k+1)} \leftarrow  & \arg\min_{\phi}
    E_{\pi_{\theta_1^{(k)}}}\Big[
    \big(r_{1}(s, a) + \gamma_{1}V_{\phi_1^{(k)}}(s') -V_{\phi}(s)  \big)^2
    \Big],\\ 
    \theta_1^{(k+1)}  \leftarrow  &\arg\max_{\theta}E_{\pi_{\theta_1^{(k)}}}\Big[ 
     \frac{\prod_{i=2}^m \Big(\pi_{\theta_i}(a|s)\Big)^{\frac{\lambda_i}{\sum_{j=2}^m \lambda_j}}}{\pi_{\theta_1^{(k)}}(a|s)}\\
    & \qquad\quad\quad\quad \times\exp\bigg(\frac{A_{1}^{(k)}}{\sum_{j=2}^m \lambda_j} \bigg)
    \log \pi_\theta(a|s) \Big].
    \end{split}
\end{equation*}
\end{small}

\textbf{Output:} the constrained policy $\pi_1$.
\end{algorithm}

\section{Offline Experiments on TripAdvisor}
\label{sec:offline}
Our code is refered to \url{https://github.com/AIDefender/TSCAC}.

\begin{table*}[h]
\caption{Performance of different algorithms on TripAdvisor}
    \centering
    \begin{tabular}{c|c|c|c|c|c|c|c|c}
    \toprule
        Algorithms &  Service & Business & Cleanliness & Check-in & Value & Rooms & Location & Overall Rating\\
        \midrule
 BC & 3.38 & -1.86 & 3.57 & -0.73 & 3.32 & 2.92 & $2.93^{*}$ & 3.92\\

 Wide\&Deep &  $3.41^{*}$ & -1.86 & $3.62^{*}$ & -0.75 & $3.36^{*}$ & 2.96 & 2.88 & 3.98\\

 RCPO & 3.40 & -1.82  & 3.61 & -0.71 & 3.34 & 2.95 & 2.91 & 3.97\\

RCPO-Multi-Critic & $3.41^{*}$ & -1.82  & $3.62^{*}$  & -0.68 & 3.35 & $2.97^{*}$ &2.87 & $3.99^{*}$\\

 Pareto & 3.36 & $-1.79^{*}$ &3.57 & $-0.62^*$ & 3.29 & 2.93& 2.86  & 3.95 \\

TSCAC & \textbf{3.43} & -1.82  & \textbf{3.64 }& -0.68  & \textbf{3.37} & \textbf{3.00} & \textbf{2.98} & \textbf{3.99}\\
\bottomrule
    \end{tabular}
\\
The results with $*$ denote the best performance among all baseline methods in each response dimension
\\the data in last row is marked by bold font when TSCAC achieves the best performance. 
\\
\label{tab:offlinetrip}
\end{table*}

In this section, we evaluate our approach on another public dataset via extensive offline learning simulations. We demonstrate the effectiveness of our approach as compared to existing baselines in both achieving the main goal and balancing the auxiliaries. 

\paragraph{Dataset} We consider a  hotel-review dataset named \emph{TripAdvisor}, which is a standard dataset for studying policy optimization in recommender system with multiple responses in \cite{chen2021reinforcement}. 
In this data, customers not only provide an \emph{overall} rating for hotels but also score hotels in multiple aspects including \emph{service}, \emph{business}, \emph{cleanliness}, \emph{check-in}, \emph{value}, \emph{rooms}, and \emph{location} \cite{alam2016joint}.
\footnote{The dataset consists of both the main objective and other responses, which can also be used to evaluate constrained policy optimization in recommender system.} Reviews provided by the same user are concatenated chronologically to form a trajectory; we filter trajectories with length smaller than $20$.
In total, we have $20277$ customers, $150$ hotels, and $257932$ reviews. 

\paragraph{MDP} A trajectory tracks a customer hotel-reviewing history.  For each  review, we have
state $s_t$: customer ID and  the last three reviewed hotel IDs as well as corresponding multi-aspect review scores; action $a_t$: currently reviewed hotel ID; reward $r_t$: a vector of eight scores the customer provided for the reviewed hotel in terms of \emph{service}, \emph{business}, \emph{cleanliness}, \emph{check-in}, \emph{value}, \emph{rooms},  \emph{location}, and \emph{overall rating}; discount factor $\gamma$: 0.99. We set the main goal to be maximizing the cumulative overall rating, and treat others as the auxiliaries.


\paragraph{Performance}

Table \ref{tab:offlinetrip} presents the results of different algorithms in terms of eight scores.  We can see that our TSCAC algorithm performs the best among all algorithms: for the main goal, TSCAC achieves the highest overall rating $3.99$; for the auxiliary goal,  TSCAC also ranks highest for $5$ out of $7$  scores (service, cleanliness, value, rooms, location). The Pareto algorithm indeed learns a Pareto optimal solution that achieves best performance on the check-in score and business score, which however does not satisfy the setting here with the main goal to optimize the overall rating. RCPO-Multi-Critic outperforms RCPO in terms of 6 scores, which validates the effect of multi-critic estimation compared with joint-critic estimation.
The RCPO-Multi-Critic algorithm achieves the same best overall score as our approach, but they sacrifice much on the others, and in particular, the location score is even lower than that from the BC algorithm.

\section{Offline Evaluation in terms of other metrics}
\label{sec: other metrics}
We also evaluate the performance of TSCAC and baseline methods in terms of the Discounted Cumulative Gain (DCG) measure, as shown in Table \ref{fig:dcg}.
\begin{table*}[htb]
    \centering
    \caption{Performance of different algorithms in terms of DCG.}
    \label{fig:dcg}
    \begin{tabular}{c|c|c|c|c|c}
    \toprule
        Algorithm  &  Click$\uparrow$ (e+2)& Like$\uparrow$ &  Comment$\uparrow$
        & Hate$\downarrow$(e-2) & WatchTime$\uparrow$(e+3)  \\
    \midrule
    BC &$4.617$ &$8.492$&$2.320$ &$8.137$ &$1.079$
    \\
    \hline
    Wide\&Deep &$4.576$ &$8.158$ &$2.274$ &$7.503$ & $1.073$
    \\
    &$-0.88\%$ & $-3.93\%$&  $-1.97\%$& $-7.79\%^{*}$ & $-0.56\%$ 
     \\
      \hline
    DeepFM   & $4.594$ &$8.439$& $2.274$ & $8.334$ & $1.083$
     \\
     &$-0.50\%$ & $-0.63\%$& $-1.97\%$& $2.43\%$& $0.41\%^{*}$
    \\
    \hline 
    RCPO &  $4.603$ & $8.537$ & $2.282$ & $8.336$ & $1.080$
    \\
    &$-0.30\%$ & $0.53\%^{*}$ & $-1.60\%$ & $2.45\%$ & $0.12\%$
    \\
     \hline 
    RCPO-Multi-Critic & $4.569$ & $8.168$ & $2.272$& $7.538$& $1.069$
    \\
    & $-1.03\%$ & $-3.81\%$ & $-2.09\%$& $-7.36\%$& $-0.88\%$
    \\ 
    \hline 
    Pareto & $4.605$ & $8.494$ & $2.304$ & $8.449$ & $1.074$
    \\
    & $-0.26\%^{*}$ & $0.02\%$ & $-0.69\%^{*}$ & $3.83\%$ & $-0.48\%$
    \\ 
    \hline 
    TSCAC & ${\bf{4.617}}$& ${\bf{8.652}}$ & ${\bf{2.378}}$& $8.036$& 
    ${1.083}$
    \\
    &{\bf{0.01\%}}& ${\bf{1.88\%}}$& ${\bf{2.49\%}}$& $-1.24\%$& 
    ${0.39\%}$
    \\ 
    \bottomrule
    \end{tabular}
        \\ $\uparrow$: higher is better; $\downarrow$: lower is better.
    \\The number in the bracket stands for the unit of this column; The number in the first row of each algorithm is the DCG score.
\\The percentage in the second row  means the performance gap between the algorithm and the BC algorithm.
\\The numbers with $*$ denote the best performance among all baseline methods in each response dimension.
\\The last row is marked by bold font when TSCAC achieves the best performance at each response dimension.
\end{table*}

\end{document}